\documentclass{article}
\usepackage{spconf,amsmath,graphicx}
\usepackage{amssymb}
\usepackage{booktabs}
\usepackage[numbers,sort&compress]{natbib}
\usepackage[pagebackref=false,breaklinks=true,letterpaper=true,colorlinks,bookmarks=false]{hyperref}
\usepackage{multirow}
\usepackage{colortbl}

\title{Bridging Unpaired Facial Photos and Sketches by Line-drawings}
\name{Meimei Shang$^{\star}$ \qquad Fei Gao$^{\star, \ddagger}$ \thanks{$\ddagger$ Corresponding author.}  \qquad Xiang Li$^{\star}$ \qquad Jingjie Zhu$^{\dagger}$ \qquad Lingna Dai$^{\star}$\thanks{\copyright 20XX IEEE. Personal use of this material is permitted. Permission from IEEE must be obtained for all other uses, in any current or future media, including reprinting/republishing this material for advertising or promotional purposes, creating new collective works, for resale or redistribution to servers or lists, or reuse of any copyrighted component of this work in other works.}}
\address{$^{\star}$ School of Computer Science and Technology, Hangzhou Dianzi University, 
Hangzhou 310018, China. \\
      $^{\dagger}$ AiSketcher Technology, Hangzhou 311215, China.}
\begin{document}
%
\maketitle
\begin{abstract}
In this paper, we propose a novel method to learn face sketch synthesis models by using unpaired data. 
Our main idea is bridging the photo domain $\mathcal{X}$ and the sketch domain $Y$ by using the line-drawing domain $\mathcal{Z}$. 
Specially, we map both photos and sketches to line-drawings by using a neural style transfer method, i.e. $F: \mathcal{X}/\mathcal{Y} \mapsto \mathcal{Z}$. 
Consequently, we obtain \textit{pseudo paired data} $(\mathcal{Z}, \mathcal{Y})$, and can learn the mapping $G:\mathcal{Z} \mapsto \mathcal{Y}$ in a supervised learning manner. 
In the inference stage, given a facial photo, we can first transfer it to a line-drawing and then to a sketch by $G \circ F$. 
Additionally, we propose a novel stroke loss for generating different types of strokes. 
Our method, termed sRender, accords well with human artists' rendering process.
Experimental results demonstrate that sRender can generate multi-style sketches, and significantly outperforms existing unpaired image-to-image translation methods. 
\end{abstract}
\begin{keywords}
Face sketch synthesis, generative adversarial networks, deep learning, image-to-image translation, neural style transfer
\end{keywords}
\section{Introduction}
\label{sec:intro}

Face sketch synthesis (FSS) aims at generating a sketchy drawing conditioned on a given facial photo \cite{Wang2013Transductive}. It has a wide range of applications in digital entertainments. 
Recently, great progresses have been made due to the success of Genearative Adversarial Networks (GANs) \cite{Goodfellow2014GAN}. Specially, researcher pay tremendous efforts to improve the quality of sketches by developing post-processing techniques \cite{wang2017bpgan}, modulating illumination variations \cite{Zhang2018IJCAI}, using ancillary information of facial composition \cite{gao2020cagan}, exploring sketch priors \cite{Chen2018Semi, Zhang2019TCYB}, and using the cycle consistency loss \cite{wang2017multgan, Zhu2019ColGAN, Zhang2019MAL}. 

Existing works formualte FSS as a \textit{paired} image-to-image (I2I) translation task \cite{Isola2017Pix2Pix}. They learn the mapping from the photo domain $\mathcal{X}$ to the sketch domain $\mathcal{Y}$ by using photo-sketch pairs in existing datasets \cite{Wang2009Face}. 
It is significant to develop methods for learning FSS models by using \textit{unpaired} photos and sketches. Such algorithms will help us simulate the style of any human artist, if only we obtain a collection of his/her sketches. 

\begin{figure}[htb]
\begin{minipage}[b]{1\linewidth}
  \centering
  \centerline{\includegraphics[width=1\linewidth]{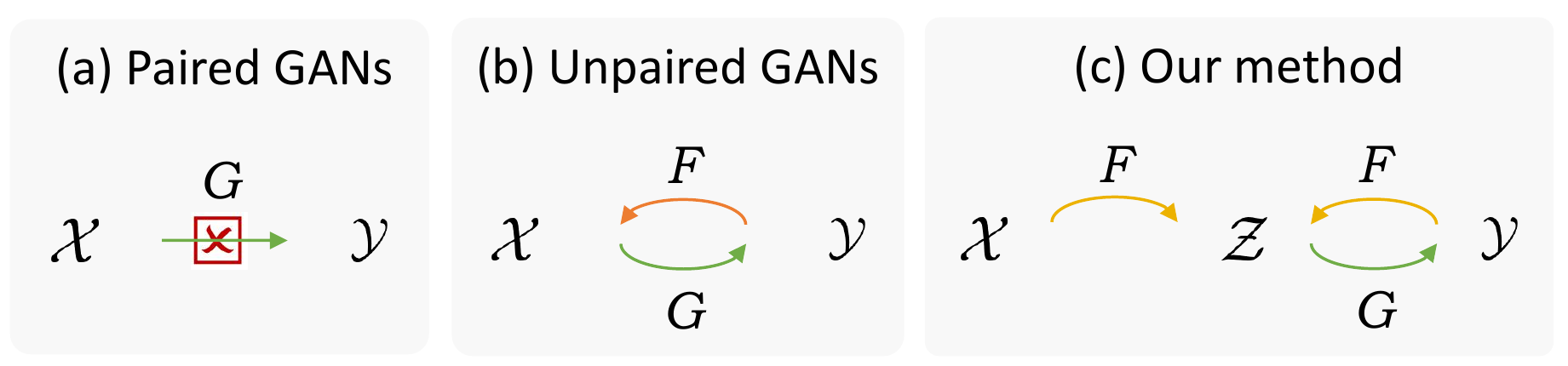}}
  \vspace{-0.4cm}
\end{minipage}
\caption{Illustration of applying paired GANs, unpaired GANs, and our method to unpaired training samples.} 
\label{fig:gans}
\end{figure}

Paired GANs cannot handle such unpaired I2I translation problem (Fig. \ref{fig:gans}a). 
To combat this challenge, researchers have proposed various unpaired GANs, by using the cycle consistency loss \cite{Zhu2017CycleGAN, Kim2020UGATIT} or learning disentangled representations in latent spaces \cite{huang2018munit, lee2018drit}. These methods simultaneously learn the mappings $G: \mathcal{X} \mapsto \mathcal{Y}$ and $F: \mathcal{Y} \mapsto \mathcal{X}$ (Fig. \ref{fig:gans}b). Although unpaired GANs perform well in various I2I translation tasks, preliminary experiments show that they fail to generate structure-consistent and stroke-realistic sketches. 

Neural style transfer (NST) is another possible solution. NST aims at transferring a content image to a target style, without changing the semantic information \cite{Gatys2015Neural}. NST methods typically need no paired examples for training. However, existing NST based FSS methods fail to generate realistic pencil-drawing strokes and textures \cite{peng2020universal, Chen2018NST}. 


To combat this challenge, in this paper, we propose to bridge \textit{unpaired} photos and sketches by line-drawings. Specially, we first map both the photo domain $\mathcal{X}$ and the sketch domain $\mathcal{Y}$ to a \textit{middle explicit} domain $\mathcal{Z}$, i.e. the line-drawings. 
To this end, we adopt a NST method, termed AiSketcher \cite{Gao2020AiSketcher}, and denotes it by $F: \mathcal{X}/\mathcal{Y} \mapsto \mathcal{Z}$. 
As a result, we obtain \textit{pseudo paired data} $(\mathcal{Z}, \mathcal{Y})$, and can learn the mapping from the line-drawing domain to the sketch domain, i.e. $G:\mathcal{Z} \mapsto \mathcal{Y}$ , in a supervised learning manner. 
The mapping from the photo domain to the sketch domain therefore becomes $G \circ F: \mathcal{X} \mapsto \mathcal{Z} \mapsto \mathcal{Y}$. 
In other words, given a facial photo $x$, we can first transfer it to a line-drawing and then to a sketch by $G(F(x))$. 

The process of our method accords well with the rendering procedure of human artists. 
When human artists draw a sketch portrait, they first use long strokes/lines to represent the outline of a given face. Afterwards, they draw small strokes and details to represent tone, space, stereoscopy, etc. 
Additionally, human artists represent diverse facial areas by using different types of strokes. We therefore propose a novel loss function to penalize the divergence between the generated and real sketches in terms of strokes. We refer to our method as \texttt{sRender}. 

%

We conduct extensive experiments on multiple styles of sketches and facial photos. Both qualitative and quantitative evaluations demonstrate that our method can generate different styles of sketches, with realistic strokes. Besides, our method significantly outperforms unpaired I2I translation methods. 

In summary, our contributions are mainly fourfold: 1) we propose a novel framework for learning sketch synthesizer from unpaired samples; 2) we propose a novel stroke loss to boost the realism of generated sketches; 3) our method accords well with human artists' rendering process; and 4) our method can generate multi-style sketches and remarkably outperforms existing unpaired I2I translation methods.
\section{Method}
\label{sec:method}

In the task of face sketch synthesis, we have the photo domain $\mathcal{X}$ and the sketch domain $\mathcal{Y}$, with \textit{unpaired} photos $\{x_i\}_{i=1}^m$ and sketches $\{y_i\}_{i=1}^n$. The goal is to learn the mapping from domain $\mathcal{X}$ to domain $\mathcal{Y}$. To handle this problem, we introduce a middle \textit{explicit} domain $\mathcal{Z}$, i.e. the line-drawings.

\begin{figure}[htb]
\begin{minipage}[b]{1.0\linewidth}
  \centering
  \centerline{\includegraphics[width=8.5cm]{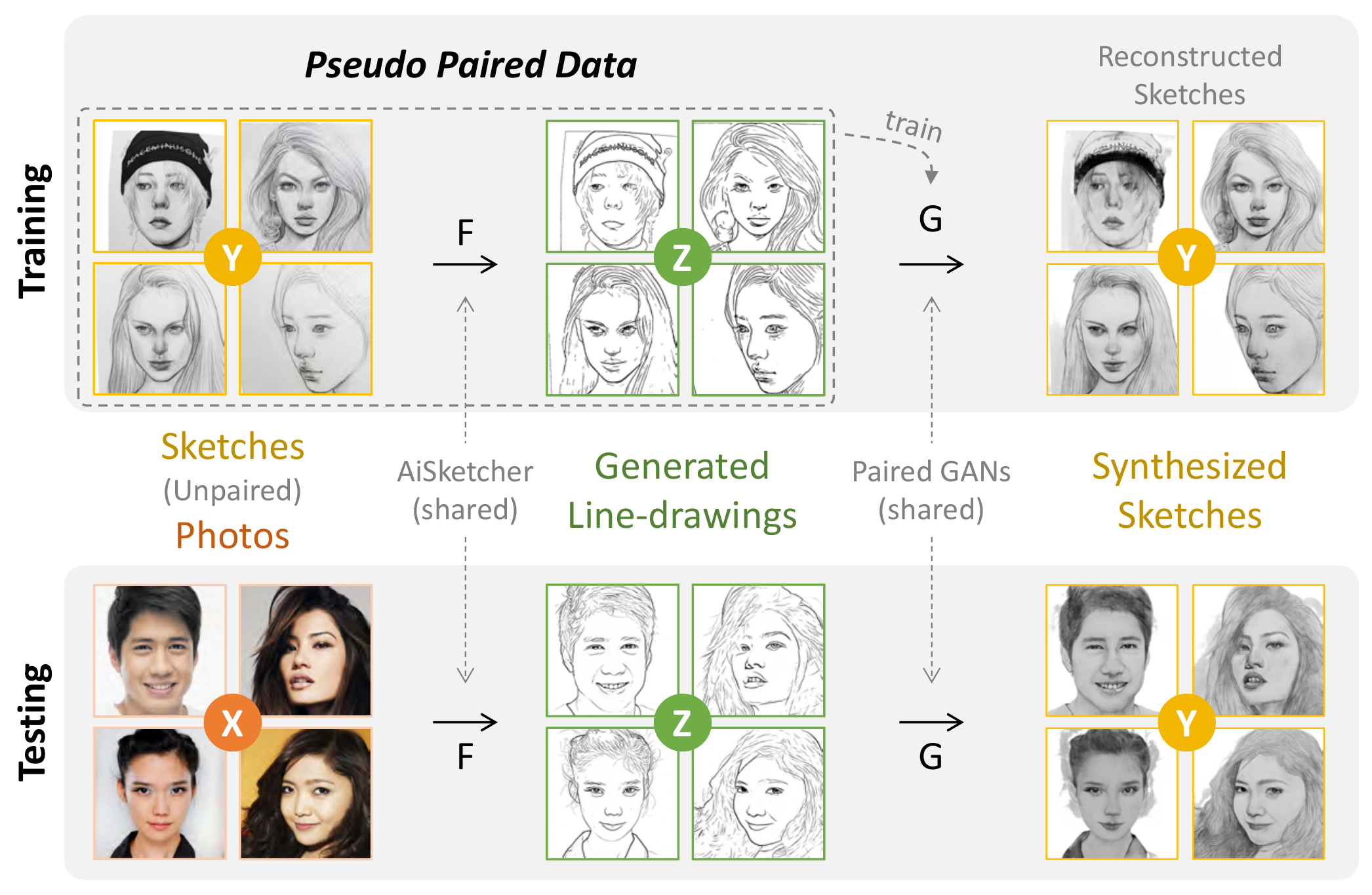}}
  \vspace{-0.4cm}
\end{minipage}
\caption{Pipeline of the proposed method.}
  \vspace{-0.4cm}
\label{fig:srender}
\end{figure}

The pipeline of our method is illustrated in Fig. \ref{fig:srender}. 
First, we map both facial photos and sketches to line-drawings by using AiSketcher \cite{Gao2020AiSketcher}: $F: \mathcal{X}/\mathcal{Y} \mapsto \mathcal{Z}$. 
In this way, we obtain pseudo paired samples: $\{(z_i, y_i)\}_{i=1}^n$ with $z_i = F(y_i)$. 
Afterwards, we learn the mapping from the line-drawing domain to the sketch domain, i.e. $G: \mathcal{Z} \mapsto \mathcal{Y}$, by using a paired GAN and such pseudo paired data. 
Finally, the mapping from the photo domain to the sketch domain becomes: $G \circ F: \mathcal{X} \mapsto \mathcal{Z} \mapsto \mathcal{Y}$. 
Details will be introduced below. 

\subsection{Line-drawing Synthesis}
\label{ssec:line}
First, we use AiSketcher \cite{Gao2020AiSketcher} as our line-drawing synthesizer, due to its remarkable performance for transferring multi-modal images to line-drawings. 
AiSketcher is an extension of AdaIN \cite{Huang2017AdaIN} with a self-consistency loss and compositional sparse loss. 
We reimplement AiSketcher exactly following \cite{Gao2020AiSketcher}, and use no paired data for training. In the testing stage, we apply the learned AiSketcher model $F$ to facial photos or sketches, and obtain the corresponding line-drawings. 
Due to the space limitation, we don't introduce AiSketcher in this paper. Please refer to \cite{Gao2020AiSketcher} for details. 

\subsection{Sketch Synthesis}
\label{ssec:gan}

We learn the mapping $G: \mathcal{Z} \mapsto \mathcal{Y}$, by using a paired GAN and the pseudo paired data $\{(z_i,y_i)\}_{i=1}^{n}$, with $z_i = F(y_i)$. 

\textbf{Network Architectures.} 
Our paired GAN includes one generator, $G$, and two discriminators, $D_k, k=1,2$. The generator contains 5 Covolutional layers, 9 residual blocks, and 5 Transposed Convolutional layers. The discriminators share the same architecture, i.e. including 5 Convolutional layers, but are fed with different scales of images. Specially, the original real and synthesized images are input into $D_1$. We downsample these images by a factor of 2 and input them into $D_2$. Such multi-scale discriminators constrain $G$ producing realistic strokes at different scales \cite{Wang2017Pix2PixHD}. 
We use ReLU and leaky ReLU in the generator and discriminators, respectively. Instance normalization is used in all networks.

\textbf{Adversarial Loss.} In the training stage, $D_k$ tries to classify the pair $(z_i, y_i)$ as positive and $(z_i, G(z_i))$ as negative. While $G$ aims to make $(z_i, G(z_i))$ classified as positive by $D_k$. The corresponding adversarial loss $\mathcal{L}_{adv}$ is expressed as:
\begin{equation}
\label{eq:ladvp}
\mathcal{L}_{adv}  = \sum_{k=1}^2 \sum_{i=1}^n \log D_k(z_i, y_i) + \log (1-D_k(z_i, G(z_i))).
\end{equation}

\textbf{Feature Matching Loss.} 
Following \cite{Wang2017Pix2PixHD}, we additionally use the feature matching loss to stabilize training. It is formulated as the L2 distance between the activations of $y_i$ and $G(z_i)$ in discriminators:
\begin{equation}
\mathcal{L}_{\mathrm{FM}} = \sum_{k=1}^2 \sum_{l=1}^5 \sum_{i=1}^n \Vert D^l_k(y_i) - D^l_k(G(z_i)) \Vert_2,
\end{equation}
where $D^l_k(\cdot)$ denotes activations on the $l$-th layer in $D_k$.



\textbf{Reconstruction Loss.} 
In our settings, given an input $z_i$, we have a real sketch $y_i$ as the target. We thus use the reconstruction loss between $G(z_i)$ and $y_i$, which is denoted by:
\begin{equation}
\mathcal{L}_{rec} = \sum_{i=1}^n \sum_j \Vert \phi^j(y_i) - \phi^j(G(z_i)) \Vert_2.
\label{eq:rec}
\end{equation}
We adopt the VGGnet pretrained for image classification as $\phi$ \cite{Johnson2016Perceptual}. $\phi^j(\cdot)$ denotes activations at the $j$-th layer of $\phi$. Preliminary experiments show that $\mathcal{L}_{rec}$ leads to more realistic textures, in contrast to the pixel-wise reconstruction loss \cite{Isola2017Pix2Pix}. 

\textbf{Stroke Loss.} 
When human artists draw a sketch, they present different facial areas by using diverse strokes. For example, they typically use long strokes to represent hairs, small strokes to represent eyebrows, stacked and light strokes to represent gradients, etc. 
To achieve this effect, we \textit{empirically} classify strokes into 7 types according to facial areas: skin, hair, boundary, eye brow, eye, clips, and ear. 
Correspondingly, we train a CNN to correctly classify the stroke type of a given patch. 
Afterwards, we fix the learned network and denote it as $\psi$. 
The stroke loss is then expressed as: 
\begin{equation}
\mathcal{L}_{str} = \sum_{i=1}^n \sum_j \Vert \psi^j(y_i) - \psi^j(G(z_i)) \Vert_2,
\label{eq:stroke}
\end{equation}
where $\psi^j(\cdot)$ denotes activations on the $j$-th layer of network $\psi$. In the implementation, $\psi$ contains an input Convolutional layer, a dense block, and an ouput Convolutional layer \cite{huang2017densenet}. Besides, we predict semantic masks of sketches by using BiSeNet \cite{yu2018bisenet}, and extract patches of the aforementioned facial areas for training $\psi$.


\textbf{Training.} 
In the training stage, we combine all the aforementioned loss functions, and optimize the generator and discriminators in an alternative manner by: 
\begin{equation}
(G^*, D_k^*) = \min_G \max_{D_k} \mathcal{L}_{adv} + \lambda_1 \mathcal{L}_{\mathrm{FM}} + \lambda_2 \mathcal{L}_{rec} + \lambda_3 \mathcal{L}_{str}.
\label{eq:opt}
\end{equation}

\textbf{Inference.} In the inference stage, the learned generator works as $G: \mathcal{Z} \mapsto \mathcal{Y}$. Recall that we have AiSketcher as $F: \mathcal{X/Y} \mapsto \mathcal{Z}$. Given a facial photo $x$, we can generate the corresponding sketch by first transferring it to a line-drawing $F(x)$ and then to a pencil-drawing by $G(F(x))$. Besides, given a real sketch $y$, we can reconstruct it by $G(F(y))$.

\section{Experiments}
\label{sec:exp}

We conduct a series of experiments to analyse the performance of our method. Details will be given below.

\begin{table*}
\centering
\caption{Quantitative evaluation of the generated croquis sketches conditioned on facial photos.}
\label{tab:comp}
\begin{tabular}{l|ccccccc}
\toprule											
 &	AdaIN \cite{Huang2017AdaIN}  & CycleGAN \cite{Zhu2017CycleGAN} & MUNIT \cite{huang2018munit} & DRIT \cite{lee2018drit} &NICE-GAN \cite{chen2020nicegan} & U-GAT-IT \cite{Kim2020UGATIT} & sRender (Ours)\\
\midrule
FID		& 	49.43 	& 45.51	& 46.35	& 42.80	& \underline{39.71}	& 48.26	& \textbf{30.35} \\
\bottomrule	
\end{tabular}													
\vspace{-0.3cm}	
\end{table*}

\subsection{Settings}
\label{ssec:setting}
For AiSketcher, we exactly follow the settings presented in \cite{Gao2020AiSketcher}. Due to space limitation, we will briefly introduce the settings about the sketch synthesis stage. Our code and results have been released at: \url{aiart.live/sRender}.

\textbf{Facial Sketches.} We collect two styles of real sketches. Specially, we download (I) 366 \textit{croquis} sketches drawn by artist \texttt{@HYEJUNG} from Instagram; and (II) 505 \textit{charcoal} sketches drawn by different artists from Web (Fig. \ref{fig:style}). 
For each style, we randomly split the corresponding sketches for training and testing in a ratio of $8:2$. 

\textbf{Facial photos.} We need no facial photos for training the sketch generator. In the testing stage, we randomly select 505 photos from the CelebA-HQ dataset \cite{CelebAMaskHQ}, and obtain the corresponding sketches by sequentially feeding them into AiSketcher $F$ and the learned sketch generator $G$.

\textbf{Preprocessing.} All images are geometrically aligned relying on two eye centers and cropped to size of $512 \times 512$. Here, we use face++ api for landmark detection. 

\textbf{Training details.} During training, each input image is rescaled to $542 \times 542$, and then a crop of size $512 \times 512$ is randomly extracted. We also use horizontal flip for data augmentation. We use the Adam solver with $\beta_1 = 0.5$, $\beta_2=0.999$, and batch size of 1. The learning rate is initially $2e-4$ for the first 100 epoch, and then linearly deceases to zero over the next 100 epochs. Besides, we set $\lambda_1 = 100$, $\lambda_2 = 10$, and $\lambda_3 = 0.002$.

\textbf{Criteria.} 
To quantitatively evaluate the quality of generated sketches, we adopt the Fr\'echet Inception Distance (FID) metric.
To alleviate the effect of facial identity, we randomly extract about 10,000 patches of size $256 \times 256$ from real and generated sketches, respectively, for calculating FIDs. 
In addition, we use the Scoot metric \cite{fan2019scoot} to measure the similarity between real and synthesised sketches. We also use the Fisherface method for face sketch recognition \cite{Zhang2019TCYB}. 

In the following, we report the FID values for both the testing sketches and the photos. While we only report the Scoot and face recognition accuracy (Acc.) on the testing sketches, because there is no real sketches for testing photos. Lower FID values, but higher Scoot and Acc. values, denote better quality of synthesised sketches.



\subsection{Results}

\begin{figure}[htb]
\begin{minipage}[b]{1.0\linewidth}
  \centering
  \centerline{\includegraphics[width=1.0\linewidth]{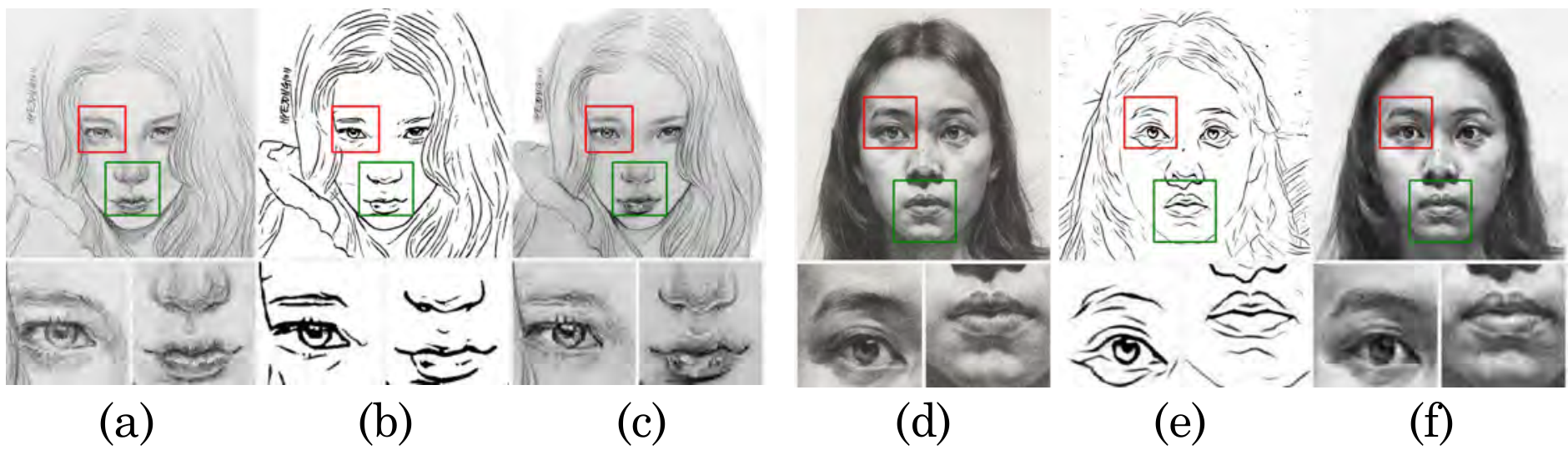}}
  \vspace{-0.4cm}
\end{minipage}
\caption{Illustration of reconstructed sketches by sRender. (a) Real croquis sketch, (b) synthesised line-drawing, (c) reconstructed sketch; (d) real charcoal sketch, (e) synthesised line-drawing, and (f) reconstructed charcoal sketch.}
\label{fig:style}
  \vspace{-0.4cm}
\end{figure}

\begin{figure}[htb]
\begin{minipage}[b]{1.0\linewidth}
  \centering
  \centerline{\includegraphics[width=1.0\linewidth]{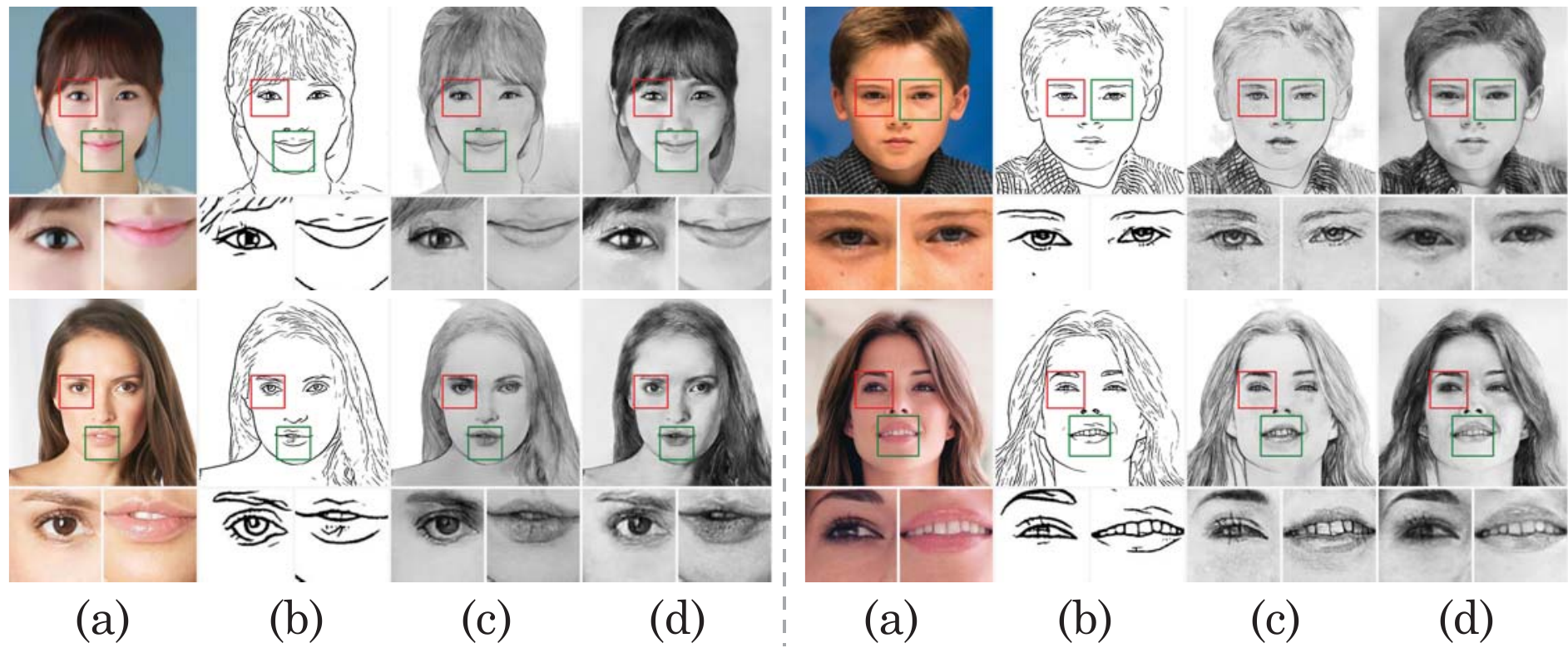}}
  \vspace{-0.4cm}
\end{minipage}
\caption{Sketches generated by sRender. (a) Input photo, (b) synthesised line-drawing, (c) generated croquis sketch, and (d) generated charcoal sketch.}
\label{fig:p2s}
  \vspace{-0.2cm}
\end{figure}

\textbf{Multi-style Sketch Synthesis.} 
In this section, we evaluate the capacity of sRender on generating multi-style sketches. 
First, we apply sRender to reconstruct real testing sketches. As shown in Fig. \ref{fig:style}, for both croquis and charcoal styles, the reconstructed sketches represent realistic strokes as the ground truths. 
Besides, sRender achieves a FID of 22.92 on the croquis sketches, and 12.30 on the charcoal sketches. 
Afterwards, we apply sRender to facial photos. As shown in Fig. \ref{fig:p2s}, the synthesis sketches preserve the content and identity of the input photos. Besides, the strokes and textures in synthesised sketches are similar to the corresponding styles shown in Fig. \ref{fig:style}a and Fig. \ref{fig:style}d. 
These observations demonstrate that sRender can generate high-quality sketches with different styles.

\begin{figure}[htb]
\begin{minipage}[b]{1.0\linewidth}
  \centering
  \centerline{\includegraphics[width=1.0\linewidth]{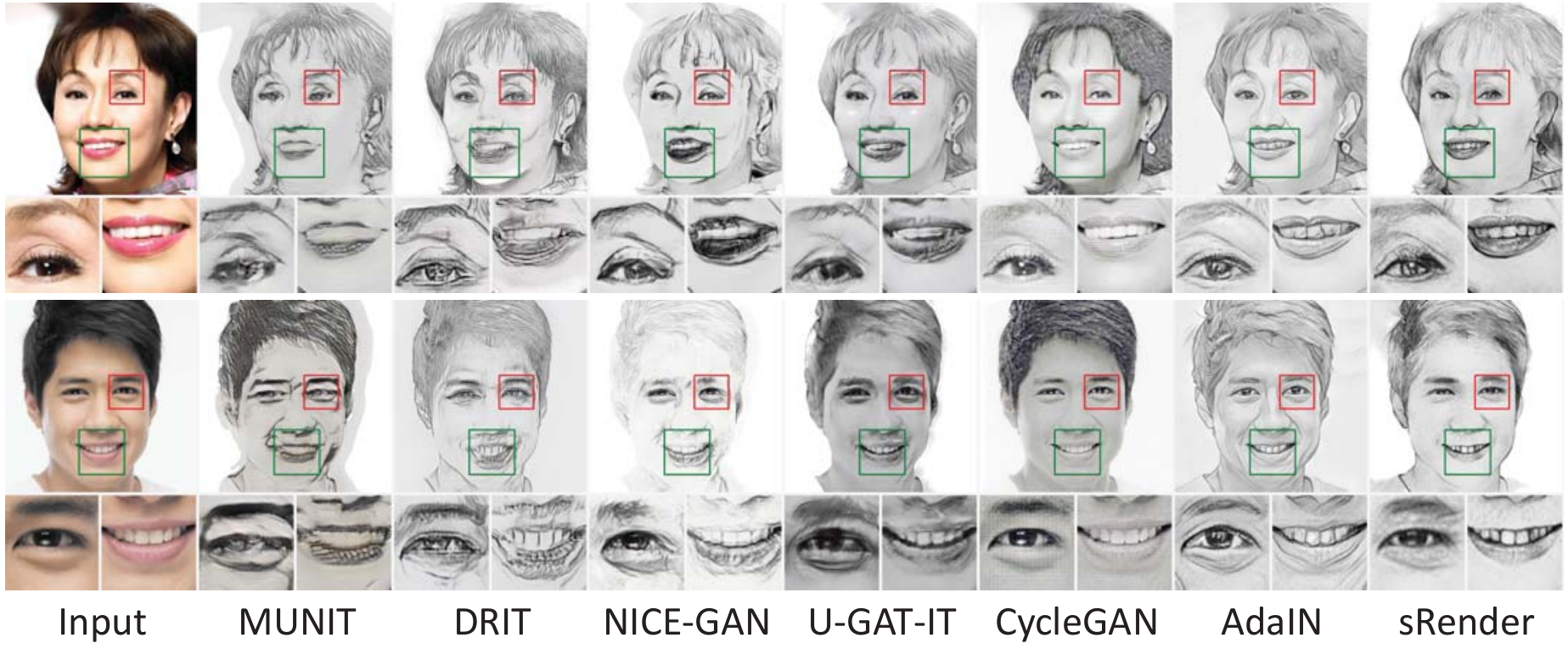}}
  \vspace{-0.4cm}
\end{minipage}
\caption{Croquis sketches generated by our sRender and unpaired I2I translation methods. (Please zoom in for details.)}
\label{fig:comp}
\end{figure}

\textbf{Comparison with SOTA.} In this part, we compare our method with various models including AdaIN \cite{Huang2017AdaIN}, CycleGAN \cite{Zhu2017CycleGAN}, MUNIT \cite{huang2018munit}, DRIT \cite{lee2018drit}, U-GAT-IT \cite{Kim2020UGATIT}, and NICE-GAN \cite{chen2020nicegan}. We train these methods by using the same unpaired data as our method. All these methods are implemented using the author's code. Since these methods learn the mapping from the photo domain to the sketch domain directly, they cannot be applied to sketches. We therefore only reports their performance on the testing photos here.

The corresponding quantitative indices are reported in Table \ref{tab:comp}, where the best indices are denoted in boldface fonts and the second best ones in underlined fonts. Obviously, our sRender obtains the best FID, which is about 9 points lower than the second best method, i.e. NICE-GAN. Such a significant superiority implies that our sRender can generate more realistic sketches than all the other methods. 

As shown in Fig. \ref{fig:comp}, the sketches generated by MUNIT, DRIT, and NICE-GAN present black inks and geometric deformations. The sketches generated by U-GAT-IT are acceptable, but still contain defects like inks. The images produced by CycleGAN are similar to grayscale photos instead of sketches. AdaIN produced visually comfortable sketches in general. However, the textures produced by AdaIN are over smooth and diverse from real pencil-drawing strokes. In contrast, sketches generated by sRender preserve the content of input photos and present realistic pencil-drawing strokes. 

Both the quantitative and qualitative comparisons demonstrate that our sRender outperforms previous methods. Besides, sRender successfully learns the mapping from the photo domain to the sketch domain, by using unpaired samples.


\textbf{Ablation Study.} 
To analyse the settings of sRender, we build two model variants by removing the stroke loss $\mathcal{L}_{str}$ (denoted by sRender w/o $\mathcal{L}_{str}$) and replacing sRender by Pix2Pix \cite{Isola2017Pix2Pix} (denoted by sRender$_\text{Pix2Pix}$), respectively. We evaluate these models on the testing sketches. 

\begin{figure}[thpb]
      \centering
      \includegraphics[width=1\linewidth]{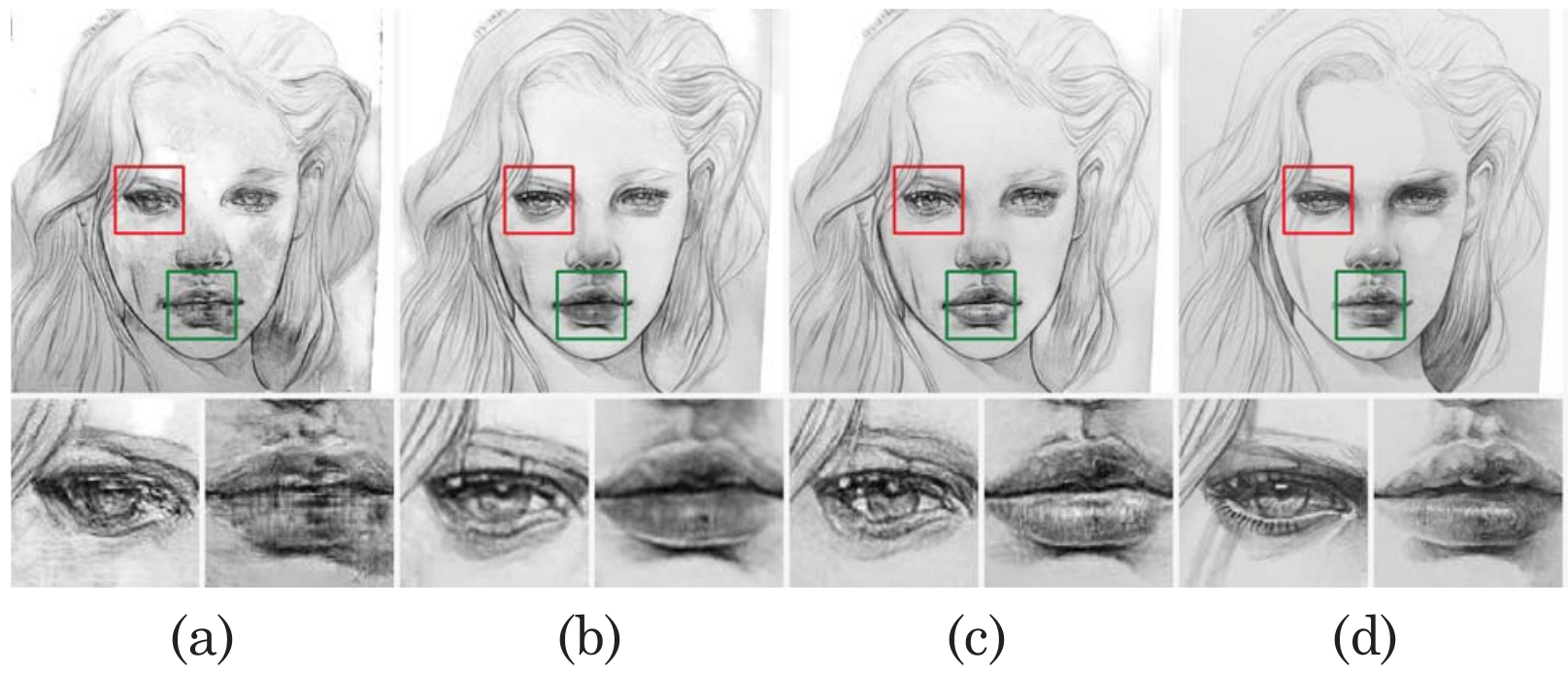}
      \vspace{-0.8cm}
      \caption{Croquis sketches generated by (a) sRender$_\text{Pix2Pix}$, (b) sRender w/o $\mathcal{L}_{str}$, (c) sRender, and (d) the ground truth.}
      \label{fig:ablation}
      \vspace{-0.6cm}
   \end{figure}

As shown in Table \ref{tab:ablation}, using $\mathcal{L}_{str}$ boosts both the quality (i.e. FID and Scoot) and the face recognition accuracy. Correspondingly, as shown in Fig. \ref{fig:ablation}, $\mathcal{L}_{str}$ make the generator produce realistic pencil-like strokes and textures. Although sRender$_\text{Pix2Pix}$ shows inferiority over sRender, it still produces high quality sketches. Such inspiring observations imply that we can extend our method to handle diverse I2I translations by exploring optimal paired GAN architectures.  

\begin{table}
\centering
\caption{Ablation study on the testing croquis sketches.}
\label{tab:ablation}
\begin{tabular*}{8.55cm}{@{\extracolsep{\fill}}l|ccc}
\toprule																
	&	sRender$_\text{Pix2Pix}$	&	sRender w/o $\mathcal{L}_{str}$	&	sRender	\\
\midrule
FID		&	37.49	&	22.97	&	\textbf{22.92}	\\
Scoot	&	0.557	&	0.570 	&	\textbf{0.587}	\\
Acc.	&	0.672	&	0.739	&	\textbf{0.750} 	\\
\bottomrule																
\end{tabular*}
\vspace{-0.5cm}	
\end{table}

%

%

\section{Conclusions}
\label{sec:con}

In this paper, we propose a novel face sketch synthesis method learned from unpaired samples. Experimental results demonstrate the remarkable capacity of our method. While there are still spaces for improving the quality of generated sketches. First, it is promising to use semantic information to guide the generators. Second, we will explore the use of facial photos in the training stage. Finally, extending the proposed method to different I2I translation tasks is another future work.

%


\bibliographystyle{IEEEbib}
\bibliography{refs}

\end{document}